\documentclass[11pt,a4paper]{article}
\usepackage[hyperref]{emnlp2018}
\usepackage{times}
\usepackage{latexsym}
\usepackage{balance}
\usepackage{url}
\usepackage{multirow}

\aclfinalcopy % Uncomment this line for the final submission

\title{UFRGS Participation on the WMT Biomedical Translation Shared Task}

\author{Felipe Soares \\
  Instituto de Informática - UFRGS \\
  Porto Alegre - RS - Brazil \\
  {\tt felipe.soares@inf.ufrgs.br} \\\And
  Karin Becker \\
  Instituto de Informática - UFRGS \\
  Porto Alegre - RS - Brazil \\
  {\tt karin.becker@inf.ufrgs.br} \\}

\date{}

\begin{document}
\maketitle
\begin{abstract}
  This paper describes the machine translation systems developed by the Universidade Federal do Rio Grande do Sul (UFRGS) team for the biomedical translation shared task. Our systems are based on statistical machine translation and neural machine translation, using the Moses and OpenNMT toolkits, respectively. We participated in four translation directions for the English/Spanish and English/Portuguese language pairs. To create our training data, we concatenated several parallel corpora, both from in-domain and out-of-domain sources, as well as terminological resources from UMLS. Our systems achieved the best BLEU scores according to the official shared task evaluation.
\end{abstract}

\section{Introduction}

In this paper, we present the system developed at the Universidade Federal do Rio Grande do Sul (UFRGS) for the Biomedical Translation shared task in the Third Conference on Machine Translation (WMT18), which consists in translating scientific texts from the biological and health domain. In this edition of the shared task, six language pairs are considered: English/Chinese, English/French, English/German, English/Portuguese, English/Romanian, and English/Spanish. \par

Our participation in this task considered the English/Portuguese and English/Spanish language pairs, with translations in both directions. For that matter, we developed two machine translation (MT) systems: one based on statistical machine translation (SMT), using Moses \cite{koehn2007moses}, and one using neural machine translation (NMT), using OpenNMT \cite{2017opennmt}. \par

This paper is structured as follows: Section 3 details the language resources used to train our translation models. Section 4 contains the description of the experimental settings of our SMT and NMT models, including the pre-processing step performed to comply with the shared task guidelines. In Section 5 we present the results and briefly discuss the main findings. Section 6 contains the conclusions and directions of future works to improve our models.

\section{Related Works}

Most of related works in biomedical machine translation used SMT models to perform automatic translation. \citet{aires2016english} developed a phrase-based SMT that differs significantly from the usual Moses toolkit, especially by not analyzing phrases at word level and adopting a translation score that is a tuned weighted average between the translation model and the language model, instead of the traditional log-linear approach. \par

\citet{costa2016talp} employed Moses SMT to perform automatic translation integrated with a neural character-based recurrent neural network for model re-ranking and bilingual word embeddings for out of vocabulary (OOV) resolution. Given the 1000-best list of SMT translations, the RNN performs a rescoring and selects the translation with the highest score. The OOV resolution module infers the word in the target language based on the bilingual word embedding trained on large monolingual corpora. Their reported results show that both approaches can improve BLEU scores, with the best results given by the combination of OOV resolution and RNN re-ranking. Similarly, \citet{ive2016limsi} also used the n-best output from Moses as input to a re-ranking model, which is based on a neural network that can handle vocabularies of arbitrary size. \par

In the last WMT biomedical translation challenge (2017) \cite{yepes2017findings}, the submission that achieved the best BLEU scores for the FR/EN language pair on the EDP dataset, in both directions, was based on NMT models developed in the University of Kyoto \cite{cromieres2016kyoto}. For the other datasets, the submission from the University of Edinburgh \cite{sennrich2017university} achieved the best BLEU scores with their NMT models based on the Nematus implementation with BPE tokenization and the use of parallel and backtranslated data.

\section{Resources}

In this section, we describe the language resources used to train both models, which are from two main types: corpora and terminological resources.

\subsection{Corpora}
We used both in-domain and general domain corpora to train our systems. For general domain data, we used the books corpus \cite{TIEDEMANN12.463}, which is available for several languages, included the ones we explored in our systems, and the JRC-Acquis \cite{TIEDEMANN12.463}. As for in-domain data, we included several different corpora: 

\begin{itemize}

\item The corpus of full-text scientific articles from Scielo \cite{SOARES18.370}, which includes articles from several scientific domains in the desired language pairs, but predominantly from biomedical and health areas. 

\item A subset of the UFAL medical corpus\footnote{\url{https://ufal.mff.cuni.cz/ufal_medical_corpus}}, containing the Medical Web Crawl data for the English/Spanish language pair.

\item The EMEA corpus \cite{TIEDEMANN12.463}, consisting of documents from the European Medicines Agency.

\item A corpus of theses and dissertations abstracts (BDTD) \cite{SOARES_Propor} from CAPES, a Brazilian governmental agency responsible for overseeing post-graduate courses. This corpus contains data only for the English/Portuguese language pair.

\item A corpus from Virtual Health Library\footnote{\url{http://bvsalud.org/}} (BVS), containing also parallel sentences for the language pairs explored in our systems.

\end{itemize}

Table \ref{table:Corpora} depicts the original number of parallel segments according to each corpora source. In Section 3.1, we detail the pre-processing steps performed on the data to comply with the task evaluation.

\begin{table}[h]
\begin{center}

\begin{tabular}{|l|c|c|}
\hline
\multicolumn{1}{|c|}{\multirow{2}{*}{\bf Corpus}} & \multicolumn{2}{c|}{\bf Sentences}     \\ \cline{2-3} 
\multicolumn{1}{|c|}{}                        & \multicolumn{1}{c|}{\bf EN/ES} & \bf EN/PT \\ \hline
Books                                         & 93,471                          & -     \\
UFAL                                   & 286,779                          & -     \\
Full-text Scielo                              & 425,631                          & 2.86M     \\
JRC-Acquis                                          & 805,757                          & 1.64M     \\
EMEA                                          & -                          & 1.08M     \\
CAPES-BDTD                                         & -                          & 950,252     \\
BVS                                           &  737,818                        & 631,946      \\ \hline
Total                                         & 2.37M                          & 7.19M     \\ \hline
\end{tabular}
\end{center}
\caption{\label{table:Corpora} Original size of individual corpora used in our experiments}
\end{table}
\subsection{Terminological Resources}

Regarding terminological resources, we extracted parallel terminologies from the Unified Medical Language System\footnote{\url{https://www.nlm.nih.gov/research/umls/}} (UMLS). For that matter, we used the MetamorphoSys application provided by U.S. National Library of Medicine (NLM) to subset the language resources for our desired language pairs. Our approach is similar to what was proposed by \citet{perezdevinaspre-labaka:2016:WMT}. \par

Once the resource was available, we imported the MRCONSO RRF file to an SQL database to split the data in a parallel format in the two language pairs. Table \ref{table:UMLS} shows the number of parallel concepts for each pair.

\begin{table}[h]
\begin{center}

\begin{tabular}{|l|c|}
\hline
\bf Language Pair & \bf Concepts \\ \hline
EN/ES         & 14,399   \\ 
EN/PT         & 26,194   \\ \hline
\end{tabular}
\end{center}
\caption{\label{table:UMLS} Number of concepts from UMLS for each language pair}
\end{table}

\section{Experimental Settings}
In this section, we detail the pre-processing steps employed as well as the architecture of the SMT and NMT systems.

\subsection{Pre-processing}
As detailed in the description of the biomedical translation task, the evaluation is based on texts extracted from Medline. Since one of our corpora, the one comprised of full-text articles from Scielo, may contain a considerable overlap with Medline data, we decided to employ a filtering step in order to avoid including such data. \par

The first step in our filter was to download metadata from Pubmed articles in Spanish and Portuguese. For that matter, we used the Ebot utility\footnote{\url{https://www.ncbi.nlm.nih.gov/Class/PowerTools/eutils/ebot/ebot.cgi}} provided by NLM using the queries \textit{POR[la]} and \textit{ESP[la]}, retrieving all results available. Once downloaded, we imported them to an SQL database which already contained the corpora metadata. To perform the filtering, we used the \textit{pii} field from Pubmed to match the Scielo unique identifiers or the title of the papers, which would match documents not from Scielo. \par

Once the documents were matched, we removed them from our database and partitioned the data in training and validation sets. Table \ref{table:Final} contains the final number of sentences for each language pair and partition.

\begin{table}[h]
\begin{center}
\begin{tabular}{|l|c|c|}
\hline
\textbf{Language} & \textbf{Train} & \textbf{Dev} \\ \hline
EN/ES             & 2.35M             & 22,670            \\
EN/PT             & 7.17M              & 24,206            \\ \hline
\end{tabular}
\end{center}
\caption{\label{table:Final} Final corpora size for each language pair}
\end{table}

\begin{table*}[t!]
\begin{center}
\begin{tabular}{|l|c|c|c|c|}
\hline
\multicolumn{1}{|c|}{\textbf{Team, Runs}} & \textbf{EN/ES} & \textbf{EN/PT} & \textbf{ES/EN} & \textbf{PT/EN} \\ \hline
UFRGS run1 (NMT)                          & 39.62          & \textbf{39.43} & 43.31          & \textbf{42.58} \\
UFRGS run2 (SMT)                          & \textbf{39.77} & \textbf{39.43} & \textbf{43.41} & \textbf{42.58} \\ \hline
TGF TALP UPC run1                         & -              & -              & 40.49          & 39.49          \\
TGF TALP UPC run2                         & -              & -              & 39.06          & 38.54          \\ \hline
UHH-DS run1                               & 31.32          & 34.92          & 36.16          & 41.84          \\
UHH-DS run2                               & 31.05          & 34.19          & 35.17          & 41.80          \\
UHH-DS run3                               & 31.33          & 34.49          & 36.05          & 41.79          \\ \hline
\end{tabular}
\end{center}
\caption{\label{table:BLEU} Official BLEU scores for the English/Spanish and English/Portuguese language pairs in both translation directions. Bold numbers indicate the best result for each direction.}
\end{table*}

\subsection{SMT System}

We used the popular Moses toolkit \cite{koehn2007moses} to train our SMT system for the two language pairs. As training parameters, we followed the Moses baseline steps\footnote{\url{http://www.statmt.org/moses/?n=moses.baseline}} to train four MT systems (i.e. one for each translation direction). \par

Regarding training, we used the Amazon AWS spot virtual machines with 24 cores and 60GB of RAM, and used parallelization as much as possible to reduce training time and the associated cost.

\subsection{NMT System}

As for the NMT system, we employed the OpenNMT toolkit \cite{2017opennmt} to train four MT systems, one for each translation direction. Tokenization was performed by the supplied OpenNMT algorithm. Regarding network parametrization, the following settings were used, while all other parameters were set as default:

\begin{itemize}
    \setlength{\itemsep}{0pt}%
    \setlength{\parskip}{0pt}%
    \item Encoder type: bidirectional recurrent neural network
    \item Decoder type: Seq2Seq with attention (default)
    \item Word vector size: 600
    \item Layers (encoder and decoder): 4
    \item RNN size: 800
    \item Batch size: 64
    \item Vocabulary size: 50000
\end{itemize}

To train our system, we used the Azure virtual machines with a single NVIDIA Tesla V100 GPU. The models with the best perplexity value were chosen as final models. During translation, OOV words were replace by their original word in the source language, all other OpenNMT options for translation were kept as default.

\section{Experimental Results}

We now detail the results achieved by our SMT and NMT systems on the official test data used in the shared task. Table \ref{table:BLEU} shows the BLEU scores \cite{papineni2002bleu} for both systems and for the submissions made by other teams. \par

Our submissions achieved the best results for all translation directions we participated, with remarkable BLEU scores for the ES/EN and PT/EN pairs. When compared to the other teams, our results presented similar behavior, with higher scores when English was the target language, which may be explained by the poor English morphosyntactic system. For the English/Spanish pair, the SMT system presented slightly better results than the NMT one, probably due to the dictionary size used in the NMT. \par

Regarding the superior results achieved, we expect that the large parallel corpora used in our experiments played an essential role. Although we did not use the provided Scielo abstracts corpus \cite{NEVES16.800}, we used a newer parallel corpus also from Scielo, but comprised of full-text articles \cite{SOARES18.370}, which overlaps with the abstracts, but contains more data. \par

In addition to the biomedical and health corpora, we employed two out-of-domain corpora that we assumed to have a similar structure to scientific texts: the books and the JRC-Acquis \cite{TIEDEMANN12.463}. We decided not to use the large Europarl corpus \cite{koehn2005europarl}, since it is comprised of speeches transcripts, which do not follow the usual structure of scientific texts.

\section{Conclusions}

We presented the UFRGS machine translation systems for the biomedical translation shared task in WMT18. For our submissions, we trained SMT and NMT systems for all four translation directions for the English/Spanish and English/Portuguese language pairs. \par

For model building, we included several corpora from biomedical and health domain, and from out-of-domain data that we considered to have similar textual structure, such as JRC-Acquis and books. Prior training, we also pre-processed our corpora to ensure, or at least minimize the risk, of including Medline data in our training set, which could produce biased models, since the evaluation was carried out on texts extracted from Medline. \par 

Our systems achieved the best results in this shared task for the translation directions we participated, which we attribute to the high quality corpora used and their size. \par

Regarding future work, we are planning on optimizing our systems by studying the following methods:

\begin{itemize}
    \item BPE tokenization: as stated by \citet{sennrich2016neural}, the use of byte pair encoding tokenization can help to tackle the issue of OOV words by using subword units. We expect that this approach can provide better results for our NMT system on biomedical data, since this domain contains terminologies that are usually based on the use of affixes.
    
    \item Backtranslation: the use of synthetic data from back-translation of monolingual proved to be able to increase NMT performance \cite{sennrich2016improving} by providing additional training data.
    
    \item Multilingual training: a study from Google \cite{johnson2017google} showed that using multilingual data when training NMT systems can improve translation performance, especially when using a many-to-one scheme (i.e. several source languages and one target language). We expect that systems trained using (ES+PT)$\rightarrow$EN, for instance, may produce better results due to the similarity between Portuguese and Spanish.
\end{itemize}

\balance
\bibliography{emnlp2018}
\bibliographystyle{acl_natbib_nourl}

\appendix

\end{document}